# Slang Context-based Inference Enhancement via Greedy Search-Guided Chain-of-Thought Prompting

Jinghan Cao*
*Department of Computer Science*
*San Francisco State University*
San Francisco, USA
jcao3@alumni.sfsu.edu
*Corresponding author

Qingyang Ren
*Department of Computer Science*
*Cornell University*
Ithaca, USA
qr23@cornell.edu

Xiangyun Chen
*Department of Biochemistry and Molecular Biology*
*Pennsylvania State University*
University Park, USA
xiangyun.amy.chen@gmail.com

Xinjin Li
*Department of Computer Science*
*Columbia University*
New York, USA
li.xinjin@columbia.edu

Haoxiang Gao
*Department of Electrical and Computer Engineering*
*Carnegie Mellon University*
Mountain View, USA
haoxiang@alumni.cmu.edu

Yu Zhao
*Independent Researcher*
New York, USA
yuzhaoqr@gmail.com

*Abstract*—Slang interpretation has been a challenging downstream task for Large Language Models (LLMs) as the expressions are inherently embedded in contextual, cultural, and linguistic frameworks. In the absence of domain-specific training data, it is difficult for LLMs to accurately interpret slang meaning based on lexical information. This paper attempts to investigate the challenges of slang inference using large LLMs and presents a greedy search-guided chain-of-thought framework for slang interpretation. Through our experiments, we conclude that the model size and temperature settings have limited impact on inference accuracy. Transformer-based models with larger active parameters do not generate higher accuracy than smaller models. Based on the results of the above empirical study, we integrate greedy search algorithms with chain-of-thought prompting for small language models to build a framework that improves the accuracy of slang interpretation. The experimental results indicate that our proposed framework demonstrates improved accuracy in slang meaning interpretation. These findings contribute to the understanding of context dependency in language models and provide a practical solution for enhancing slang comprehension through a structured reasoning prompting framework.

## I. Introduction

Large Language Models (LLMs) have been extensively investigated and applied across various downstream tasks, including content writing [1], summarization [2], information extraction [3], content understanding [4], and classification [5]. Trained on vast amounts of textual data, these models demonstrate proficiency in capturing complex language patterns and contextual relationships.

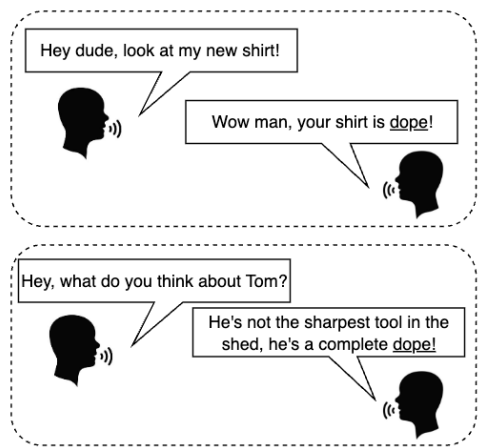

Fig. 1. Context-Dependent Slang Interpretation

However, interpretation of informal languages, particularly slang expressions remains a significant challenge. While language models can easily infer the meaning if the slang datasets are used for pre-training, they may fail to correctly infer the meaning of unseen slang, even when context is provided. The rapid evolution of slang through social media and generational changes coupled with the substantial costs of model training, makes it impractical to continuously update models with emerging slang data. In this paper, we propose an enhanced chain-of-thought prompting approach that incorporates greedy search algorithms to improve models' reasoning capabilities for slang interpretation and meaning inference.

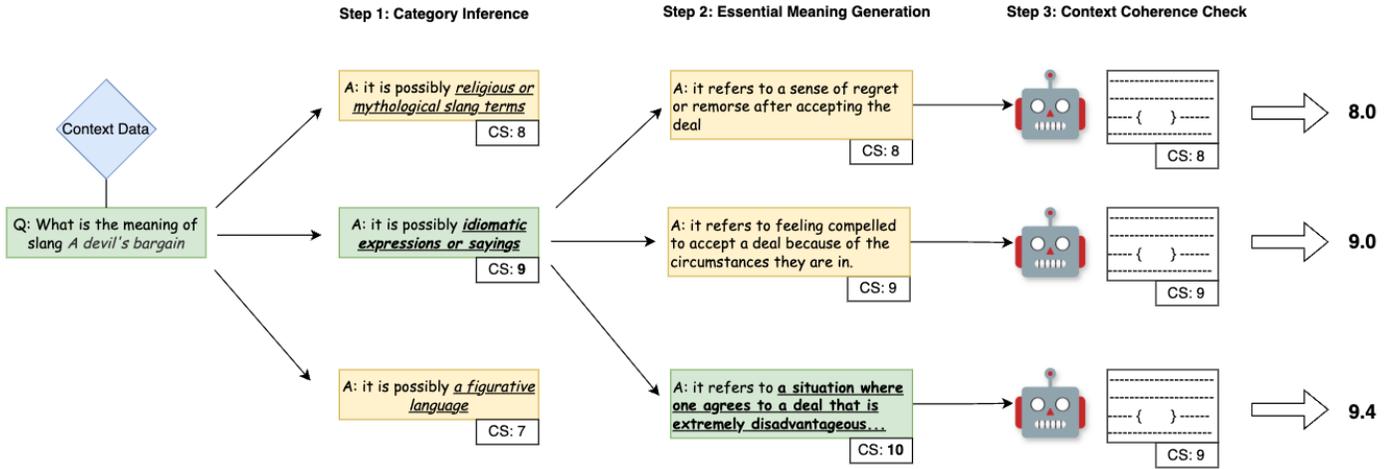

Fig. 2. Greedy Search-Guided Chain-of-Thought

This study comprises two major components. First, we conducted a comparative analysis across different models to investigate the impact of temperature settings and model size on slang interpretation accuracy. Based on this empirical finding, we then proposed a greedy search-guided chain-of-thought framework and evaluated its performance in comparison with the conventional Input-Output (IO) prompting approach.

The primary contributions of this paper are as follows:

- We empirically evaluate two meta parameters (model size and temperature) in prompting approach to perform slang meaning inference. Our experimental results demonstrate that increasing model size yields limited improvements, while higher temperature settings may adversely affect task accuracy.

- We propose a novel application of chain-of-thought approach integrated with greedy search algorithm to enhance language models' inference capabilities. This approach extends the application of chain-of-thought prompting beyond its traditional use in arithmetic reasoning tasks, demonstrating its effectiveness in improving models' general inference abilities.

## II. RELATED WORK

Slang, as a form of informal language, has been rapidly evolving and adapting to the widespread social media and digital communication platforms [6]. However, in the field of natural language processing, slang-related research is still insufficient. Unlike mathematical reasoning tasks that converge to definitive solutions, slang interpretation is inherently dynamic and context-dependent, varying across cultural, linguistic, and regional dimensions [7][8][9]. Language models trained on historical corpus data have limited inference capability to "guess" a given slang term's primary meaning. For example, as shown in Fig. 1, the term *Dope* has multiple meanings in different contexts. It can be used to express *something is pretty cool*, but it also refers to *stupid person* [10]. Given the dynamic nature of slang expressions, existing research can be classified into two major areas:

**Slang Detection.** To explore if the trained models are capable of identifying slang terms in everyday communication, researchers have made efforts to propose novel model training strategies using various data sources and to introduce benchmarks to assess models' capability in performing this task. For example, Sun et al. [11] constructed a slang dataset composed of dialogues extracted from English movies as the benchmark and enhanced the LLMs' ability to detect slang expressions in a sentence and infer their sources by applying pretrained Transformer models and fine-tuned GPT models. In addition to the innovation in the comparative analysis on models, Pei et al. [12] proposed a novel model structure by building bidirectional long short-term memory (Bi-LSTM) model with multilayer perceptron [13] to detect if slang exists at both sentence-level and token-level. Similarly, Seki and Liu [14] introduced a two-layered bidirectional LSTM with a hierarchical multi-task learning approach to detect Internet slang in Japanese.

**Slang Interpretation.** As early as 2017, Ni and Wang [15] collected large amounts of slang data from Urban Dictionary and utilized a dual encoder structure in an LSTM model to generate slang explanations. With the emergence of LLMs, Wuraola et al. [16] conducted research on employing transformer-based model in identifying the hidden emotional nuances from climate-related tweets and highlighted the shortcomings of language models in cultural insensitivity. Similarly, Mei et al. [17] extended the context of slang interpretation to daily communication scenarios and introduced a benchmark to evaluate existing language models' comprehension of new phrases given the colloquial context.

By reviewing the relevant research on slang detection and interpretation, a notable gap exists in approaches to augment a language model's intricate comprehension ability in performing downstream tasks. In this study, as illustrated in Fig. 2, we propose a novel chain-of-thought prompting approach enhanced by greedy decoding algorithms to generate optimal context-aware explanations for novel slang expressions.

## III. METHODOLOGY

### A. Problem Formulation

Each slang record consists of word $S_{word}$, ground truth meaning $M_{origin}$ and usage example $C_{example}$. The objective is to generate the most-relevant slang meaning $M_{gen}$ that has the highest sematic similarity with $M_{origin}$. Formally we break down this process into two main steps.

Firstly, given the input $S_{word}$ and $C_{example}$, the best guessed slang meaning $M_{gen}$ is generated as the output

$$M_{gen} = f(S_{word}, C_{example}; P_{cot}) \quad (1)$$

where $P_{cot}$ represents the chain-of-thought prompts.

Secondly, given input x as an element of $M_{gen}$ and the ground truth meaning $M_{origin}$, the output S represents the generated meaning with the highest semantic similarity score relative to $M_{origin}$

$$S = argmax_{x \in M_{gen}} sim(x, M_{origin}) \quad (2)$$

where $sim(x, M_{origin})$ denotes the semantic similarity between generated meaning and the ground truth meaning.

### B. Greedy Search-Guided Chain-of-Thought

Chain-of-thought strategy has been proven effective in improving the ability of LLMs to perform complex reasoning tasks [18]. Inspired by the self-consistent chain-of-thought strategy [19] and tree-of-thought strategy [20], we have developed a specialized chain-of-thought workflow for inferring slang meanings. As indicated in Fig. 3, we introduce confidence score and employ a greedy algorithm to identify the most probable explanation for slang expressions. Unlike the conventional Input-Output (IO) single chain, our method expands the number of candidate thoughts to W with a depth of D, where W represents the number of possible candidates at each step and D represents the total number of steps in the chain. Both W and D are task-dependent parameters. In our experiments, we set both W and D to 3, aligning with the designed prompting steps..

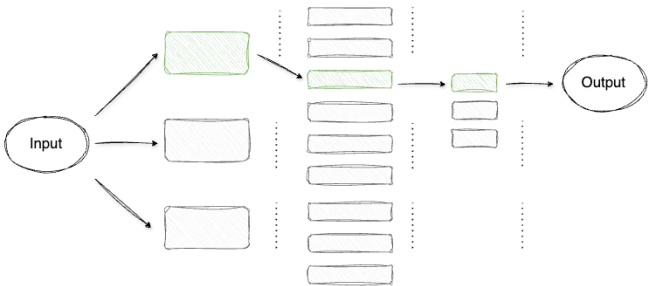

Fig. 3. Greedy Algorithm in Chain-of-Thought

### C. Step-by-step Prompting

We break the chain-of-thinking process into a three-stage inference framework.

**Category**. An initial prompt is designed to help the model identify possible categories of slang based on given usage examples. As described in Algorithm 1 following principal of greedy algorithm, we select the inferred category with highest confidence score.

---
**Algorithm 1** Category Inference
**Inputs**: *Context*: the usage example; *Slang*: the target slang term; K: total number of candidates to be generated.
*Prompt:* prompt to instruct LLM to generate categories.
**Output**: *maxTuple*: the category with the highest score
1: *maxTuple* ← *(null, 0)*
2: *thoughts* ← *LLMInfer (Context, Slang, K, Prompt)*
3: **for** *(category, score)* in thoughts **do**
4:   **if** *maxTuple* [1] < *score* **then**
5:     *maxTuple* ← *(category, score)*
6:   **end if**
7: **end for**
8: **return** *maxTuple*

---

**Essential Meaning**. In the second prompting iteration, we instruct the LLM to generate possible primary meanings based on the previously inferred category. The process is similar to Algorithm 1 except for an additional input parameter representing the inferred category. The output is a list of candidate meanings for the target slang.

---
**Algorithm 2** Essential Meaning Generation
**Inputs**: C*ategory*: the selected category; *Slang*: the target slang term; K: total number of candidates to be generated.
*Prompt:* prompt to instruct LLM to generate meanings.
**Output**: *meaningList*: a list of generated meanings with confidence score
1: *meaningList* ← *[]*
2: *meaningList* ← *LLMInfer (Category, Slang, K, Prompt)*
3: **return** *meaningList*

---

**Compatibility**. In the final step, we select the slang interpretation with the highest confidence score and prompt the LLM to evaluate whether this meaning aligns with the original context and tone. A heuristic weighting mechanism is applied in calculating the final confidence score.

---
**Algorithm 3** Context Coherence Check
**Inputs**: *Context*: the usage example; *meaningList*: a list of generated meanings with confidence score. *Prompt:* prompt to instruct LLM to check compatibility in original context
**Output**: *selectedMeaning*: the selected meaning
1: selectedMeaning ← *null*
2: *finalScore* ←*0*
3*: selectedMeaning* ← *null*
4: **for** *(meaning, score)* in meaningList **do**
5:   *confidenceScore* ← *LLMInfer (Context, meaning, Prompt)*
6:   **if** *finalScore* < *(confidenceScore \*0.6 + score \* 0.4)*
7:     *finalScore* ← *confidenceScore \*0.6 + score \* 0.4*
8*:*     *selectedMeaning* ← *meaning*
9:   **end if**
10**: end for**
11: **return** *selectedMeaning*

## IV. EXPERIMENT AND ANALYSIS

We conducted two sets of experiments. In the first experiment, we evaluated the inference capabilities across different model sizes and temperature settings using standard IO prompting. Based on these findings, we then proceed with the second experiment implementing the proposed greedy chain-of-thought prompting strategy and comparing its performance with the conventional CoT approach.

### A. Experimental Setting

**Configuration.** In the first experiment, we evaluated multiple language models, including GTP-4o, Qwen2.5-72B, DeepSeek-V3 and their smaller versions including GPT-4o-mini, Qwen2-7B-Instruct and DeepSeek-R1-Distill-Llama-8B. For evaluation, we randomly selected and preprocessed 1,200 high-quality slang records for testing across GPT, Qwen, and DeepSeek models. Additionally, we investigated the impact of temperature settings using Qwen2-7B-Instruct as the baseline model, testing temperatures of 0.1, 0.3, 0.5, and 0.7 across 500 slang records.

In the second experiment, we compared standard IO promoting with our proposed chain-of-thought approach incorporating greedy search using Qwen2-7B-Instruct as the base model. The temperature was set to 0.3 and tested against 1200 random processed slang records.

For both experiments, we utilized OpenAI APIs for GPT series models, while Qwen and DeepSeek models were locally hosted and evaluated on a cluster of 8 NVIDIA A100 GPUs with a total of 320GB GPU memory.

**Dataset Preprocessing.** The raw dataset comprises highly-voted slang entries extracted from Urban Dictionary [10], with each record containing three attributes: slang word, meaning, and usage example. Given the variable quality of Urban Dictionary records, we employed GPT-4o to filter out mismatched entries and standardize both ground truth meanings and usage examples. For usage examples specifically, we reformatted the raw content into structured dialogues to provide clearer context (Fig. 4).

TABLE I. IMPACT OF MODEL SIZE ON INFERENCE

| Models | ROUGE-L | | | SimCSE |
|---|---|---|---|---|
| | F1 | Precision | Recall | |
| GPT-4o | 0.225 | 0.149 | 0.171 | **0.736** |
| GTP-4o-mini | **0.299** | **0.166** | **0.250** | 0.727 |
| Qwen2.5-72B | 0.123 | 0.250 | 0.199 | **0.715** |
| Qwen2-7B-Instruct | **0.170** | **0.322** | **0.222** | 0.696 |
| DeepSeek-V3 | **0.235** | 0.166 | **0.300** | **0.726** |
| DeepSeek-R1-Distill-Llama-8B | 0.166 | 0.111 | 0.222 | 0.696 |

### B. Metrics

**ROUGE-L.** To evaluate the similarity between inferred meanings and ground-truth references, we employ ROUGE-L [21], which computes sentence-level content similarity through F1 score, precision, and recall metrics.

**SimCSE.** This sentence-level embedding benchmark [22] demonstrates superior performance in semantic textual similarity tasks using supervised BERT-based models.

### C. Experimental Results

*a) Limited Impact from Model Size and Temperature Setting:* Based on experimental results comparing multiple model sizes, we observe that larger models do not necessarily demonstrate superior performance, as shown in Table I. Interestingly, smaller models such as GPT-4o-mini and Qwen2-7B-Instruct achieve higher F1 scores despite lower SimCSE results.

Then we conducted experiments with incremental temperature settings on Qwen2-7B-Instruct and DeepSeek-R1-Distill-Llama-8B. As shown in Fig. 5, our experiments reveal that higher temperature settings do not correlate with improved slang inference performance. This finding aligns with prior research [23], which established that temperature variations have minimal impact on LLMs' problem-solving capabilities.

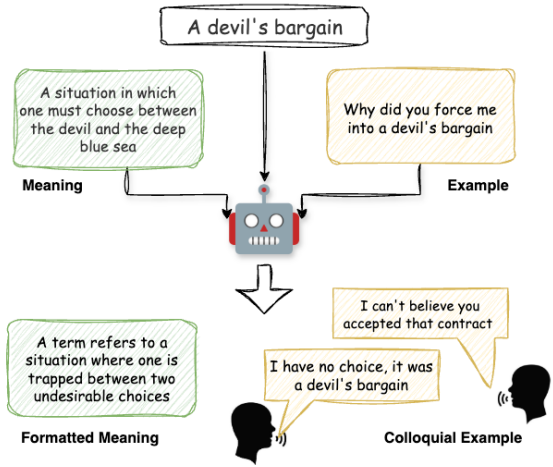

Fig. 4. Rephrase the original meaning and example

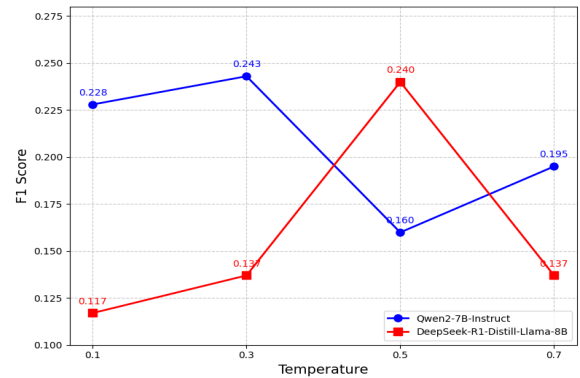

Fig. 5. Temperature Impact on Inference Ability

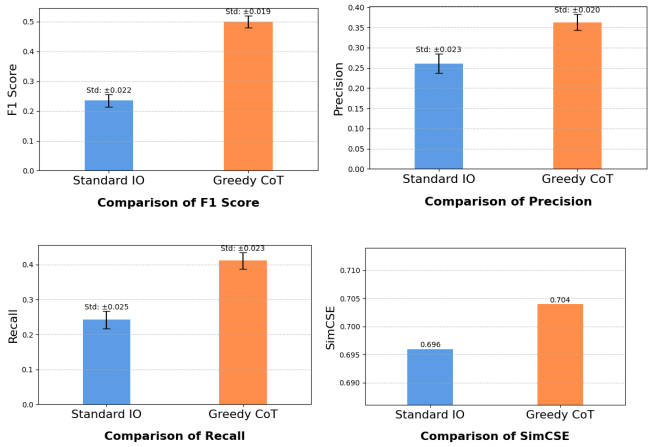

Fig. 6. Performance Comparison between standard IO Prompt and Greedy Search-Guided CoT Prompt

*b) Improved Accuracy with Greedy Search-Guided Chain-of-Thought:* Given the superior overall performance of Qwen2-7B-Instruct with the temperature set to 0.3, we proceeded to the second experiment implementing our proposed approach.

As illustrated in Fig. 6, the comparative analysis revealed significant performance differences. The standard Input-Output approach achieved an F1 score of 0.235, precision of 0.261, recall of 0.242, and SimCSE score of 0.696. In contrast, our proposed greedy chain-of-thought method demonstrated superior performance across all metrics, achieving an F1 score of 0.500, precision of 0.363, recall of 0.411, and SimCSE score of 0.704.

The experimental results highlight the effectiveness of greedy search-guided chain-of-thought prompts. Unlike traditional reasoning tasks, slang inference presents unique challenges in evaluating generated candidates at each step, as the LLM lacks access to the ground-truth meaning. Our approach addresses this by expanding the number of generated candidates heuristically, setting the candidate count to three for initial performance verification.

Analysis of experiment logs revealed that the LLM's first response is not consistently the most confident. This can be attributed to the non-deterministic nature of LLMs and the complex token sampling mechanisms that generate varying results. Moreover, confidence score assignment operates as an independent process, influenced by input parameters and meta-configuration settings.

## V. Conclusion

This paper presents a novel approach to enhance slang interpretation capabilities in language models through a greedy search-guided chain-of-thought prompting framework. Our research reveals several significant findings:

First, our empirical analysis revealed that model size and temperature settings have limited impact on slang interpretation accuracy. Contrary to common assumptions, larger models did not consistently outperform their smaller counterparts, and higher temperature settings did not lead to improved inference capabilities.

Second, our proposed framework demonstrated substantial improvements over standard Input-Output prompting, achieving higher scores across all evaluation metrics. This improvement can be attributed to the framework's structured approach to reasoning and its ability to evaluate multiple candidate interpretations at each step.

The success of our approach suggests that breaking down slang interpretation into discrete reasoning steps enables more accurate and contextually appropriate interpretations. This finding has important implications for improving language models' handling of informal language without requiring additional training or model parameter modifications.

Future work will explore generalization of this approach to be applied in other inference tasks and develop more sophisticated confidence scoring mechanisms for candidate evaluation.

## APPENDIX

This section outlines the step-by-step sample prompts to instruct the model to generate expected results.

```
System Prompt
    You are an expert in urban slang and internet language

User Prompt
    Given the context:
    Example usage: {context}

    Please explain the following word:
    Word: {slang_word}

    Based on the Word morphology and Example usage, infer which category this slang term most likely belongs to, provide 3 possible options with confidence score. You must return a JSON response with exactly this format
    {{
        "Your_First_Thought": "your first thought"
        "Your_First_Thought_Score": [0-10]

        "Your_Second_Thought": "your second thought"
        "Your_Second_Thought_Score": [0-10]

        "Your_Third_Thought": "your third thought"
        "Your_Third_Thought_Score": [0-10]
    }}
```

Fig.7. Sample Prompt to Infer Category

The first step in the prompt chain requires the model to infer the top-K (K=3) possible categories before generating any specific meanings.

```
System Prompt
    You are an expert in urban slang and internet language

User Prompt
    Given the original context:
    Example usage: {original_context}

    And the most likely inferred category
    Inferred category: {inferred_category}

    Based on Inferred category, what is the essential meaning of this term ? Provide 3 possible options with confidence score.
    You must return a JSON response with exactly this format
    {{
        "Your_First_Thought": "your first thought"
        "Your_First_Thought_Score": [0-10]

        "Your_Second_Thought": "your second thought"
        "Your_Second_Thought_Score": [0-10]

        "Your_Third_Thought": "your third thought"
        "Your_Third_Thought_Score": [0-10]
    }}
```

Fig. 8. Sample Prompt to Generate Possible Meanings

The LLM takes the most likely category as input for the next prompt. Based on this input, it generates three possible meanings and assigns confidence scores to each.

```
System Prompt
    You are an expert in urban slang and internet language

User Prompt
    Given the original context:
    Example usage: {original_context}

    And the most likely inferred category
    Inferred category: {inferred_category}

    And the most likely inferred meaning
    Inferred meaning: {inferred_meaning}

    Evaluate if the inferred meaning fit into the original context and provide your confidence score in terms of the compatibility.
    {{
        "Your_Confidence_Score: [0-10]
    }}
```

Fig. 9. Sample Prompt to Provide Confidence Score for Selected Meaning

In the final step, the model evaluates the compatibility between the original context, inferred categories, and derived meanings to generate a confidence score.